\documentclass{article}
\usepackage[utf8]{inputenc}
\usepackage[english]{babel}
\usepackage{acronym}
\usepackage{amsmath,amsfonts,amssymb}
\usepackage{graphicx}
\graphicspath{ {./images/} }
\usepackage{hyperref}
\usepackage{algorithm}
\usepackage{algpseudocode}

\usepackage{todonotes}

\usepackage{amsthm}
\newtheorem{theorem}{Theorem}[section]

\newtheorem{definition}[theorem]{Definition}
\newif\ifuseboldmathops
\newif\ifuseittextabbrevs
\useboldmathopstrue   

\ifuseittextabbrevs

\else

\fi

\ifuseboldmathops

\else

\fi

\ifuseboldmathops

\else

\fi

\ifuseboldmathops

\else

\fi

\ifuseboldmathops

\else

\fi

\acrodef{mdp}[MDP]{Markov Decision Process}

\newcommand{\isep}{\mathrel{{.}\,{.}}\nobreak}

\title{Improved Vehicle Maneuver Prediction using Game Theoretic priors}
\author{Nishant Doshi}
\date{Senior Software Engineer, Apex.AI}

\begin{document}

\maketitle

\noindent{\itshape Note: This project was completed in 2021 at Traxen Inc. and is not associated with the author’s current position. \par}

\section{Overview}

Conventional maneuver prediction methods use some sort of classification model on temporal trajectory data to predict behavior of agents over a set time horizon. Despite of having the best precision and recall, these models cannot predict a lane change accurately unless they incorporate information about the entire scene. Level-k game theory can leverage the human-like hierarchical reasoning to come up with the most rational decisions each agent can make in a group. This can be leveraged to model interactions between different vehicles in presence of each other and hence compute the most rational decisions each agent would make. The result of game theoretic evaluation can be used as a “prior “ or combined with a traditional motion-based classification model to achieve more accurate predictions. The proposed approach assumes that the states of the vehicles around the target lead vehicle are known. The module will output the most rational maneuver prediction of the target vehicle based on an online optimization solution. These predictions are instrumental in decision making systems like Adaptive Cruise Control (ACC) or Traxen's iQ-Cruise further improving the resulting fuel savings. 

\section{The problem statement}
To find the most likely maneuver of a nearby vehicle for a finite horizon, based on the observed level of reasoning of the drivers driving vehicles around it.  

\section{Definitions}
The maneuver prediction module gets processed information from Perception module in the form of relative vehicle distances and velocities. It also has access to semantic information like road geometry, number of lanes, upcoming features (exits, merges, forks) through HD maps data.

\begin{definition}[Vehicle State]
This consists of local positions with respect to host vehicle, and absolute velocities and accelerations of the vehicle being considered. 
\end{definition}

\begin{definition}[Joint State]
We define the joint state of the scene as $\{s_1,s_2,\dots,s_m\}$ where m is the number of vehicles in the scene.\\
\end{definition}

\begin{definition}[Maneuver]
A maneuver is indicated by $D^i$ which is essentially a function outputting $s_t$ of a vehicle $i$ at any time t after the maneuver starts. For a finite horizon, a sequence of maneuvers is defined as a policy which is indicated by $\Pi$ for target lead vehicle and for other vehicles in the scene, $\hat{\Pi}_k^i$ where k is the level of the vehicle in decision-making hierarchy.
\end{definition}

\section{Vehicle Maneuver Models}
\begin{figure}[h]
\centering
\includegraphics[width=0.6\textwidth]{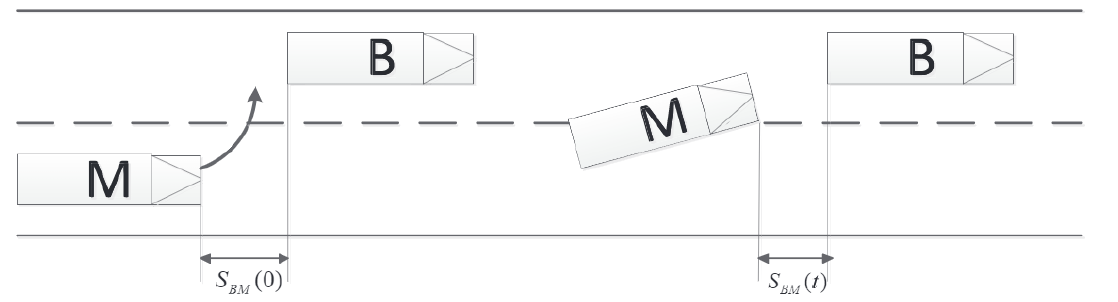}
\caption{Example of a lane change}
\end{figure}

Figure 1 is an example of a classic lane change process. Here, Vehicle M is changing lanes and then follows vehicle B.
In the process of the vehicle B changing lane, the driver's attention moves towards the motion of Vehicle B. This model is simple, practical and widely used in simulation environments and its expression is:

\begin{equation}
    d_{ref} = c_1 + c_2.v(t) 
\end{equation}
\begin{equation}
    d_{err} = d_{ra} - d_{ref}
\end{equation}
\begin{equation}
    a_{re{f\_d}} = c_d.d_{err} + c_p(V_b(t)-V_m(t))
\end{equation}
The parameters are described in the figure below.
\begin{figure}[h]
\centering
\includegraphics[width=1\textwidth]{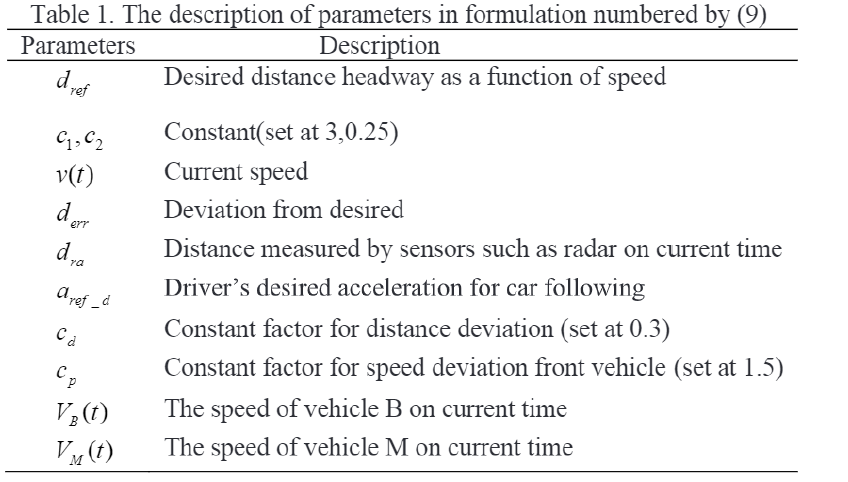}
\caption{Example of a lane change}
\end{figure}

\section{Formulation}

\paragraph{Action Set}
The action set for each vehicle consists of discrete maneuvers:
\begin{itemize}
    \item Accelerate
    \item Decelerate
    \item Left Lane Change
    \item Right Lane Change
    \item No Action
\end{itemize}
Each of these discrete actions has a predefined parametrized maneuver model (as explained in Section 3) associated with it. 

\paragraph{Reward Function}
Defining multi objective reward function:\\
\[
C(s)=w_1R_c+w_2R_s+w_3R_o+w_4R_f\:\:\:\:\:\dots[1]
\]
where,
\begin{itemize}
    \item $C(s)$ is the reward for joint state s
    \item $\{w_1,w_2,w_3,w_4\}$ are the weights $w_i\in\mathbb{R}$
    \item $R_c$ is the collision reward (0 if no collision else -1)
    \item $R_s$ is the safety reward (0 if vehicle safety envelope do not overlap else -1). Safety envelope can be assumed as a rectangular box 1.25 times the dimension along length and width. 
    \item $R_o$ is the road rule violation reward (-1 if vehicle goes into wrong lane or curb)
    \item  $R_f$ is objective-based reward (in this case, more reward if the joint state has more average velocity) 
\end{itemize}

\paragraph{Cost over horizon}
We consider finite horizon objective function for a horizon of 5 seconds which is 100 steps of 50ms. the cumulative cost is:
\[ 
C(\::|\:\hat{\Pi}_k^i)=\sum_{t=0}^{100} \gamma^t C(s_t)\:\:\:\:\:\dots[2]
\]
where $\gamma$ is the discount factor.

\paragraph{Level-k Game Theory}
Lets consider two agents A(Red car) and B(Blue car). Agent A want to do a lane change to right. 
\begin{figure}[h]
\centering
\includegraphics[width=0.6\textwidth]{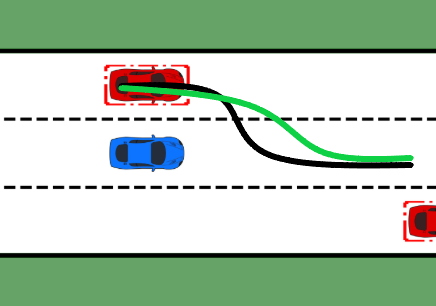}
\caption{A common scenario}
\end{figure}
Level-k game theory uses hierarchical reasoning to make decision. Let's say agent A is a level-0 agent. This means it would act aggressively without caring about what agent B might do. In this case it would accelerate and do a sharp cut-in following the black trajectory. A level-1 agent is generally more cautious i.e. it would consider the possibility that agent B might accelerate or perform some other aggressive maneuver and makes a more conservative lane change (green).

\paragraph{Level-k policies}
Level-k games is a framework for hierarchical reasoning where an agent will make a level-k decision assuming that all other agents are following level (k-1) policies. In that sense, level-0 agent exhibits a highly aggressive behavior as it assumes the other agents are stationary. A level-1 agent would assume that all the other agents are level-0 and make rational decision based on that.
It has been experimentally inferred that for most human drivers, $k\in[0,2]$.   
This formulation involves maintaining and updating a modeling of  level of decision-making of other vehicles i.e. k by the target lead vehicle. This means that the target lead vehicle would maintain a probability for the other vehicles behaving as level-i agents where $i\in[0,2]$.
Let $P$ be the probability distribution over $k\in\{0,1,2\}$.
Equation [2] is extended as follows:
\[
C(\::|P,\hat{\Pi})=\sum_{k=0}^{2} P(k) C(\::|\:\hat{\Pi}_k^i)\:\:\:\:\:\dots[3]
\]
Here,\[
\hat{\Pi}=\{\dots,\hat{\Pi^i_k},\dots\} \: \forall i\in[1,m],\:k\in[0,2]\}
\]
\paragraph{Optimization Problem}
We need to select a set of maneuvers for the target vehicle given the other vehicles following $\hat{\Pi}^i_k$ such that the cumulative reward is maximized i.e. 
\[
\Pi^j=\arg\max_{\Pi^j}\:C(\::|\:P,\hat{\Pi})\:\:\:\:\:\dots[4]
\]
where $j\in\{1,\dots,100\}$
subject to constraints
\[
s_t^i=D_j(s_{t_j}^i,t-t_j)\: \forall\:i\in[1,\dots,m] \:\:\:\:\:\dots[5]
\]
Here,\[
j=
\begin{cases}
    \Pi_j(s_t),& \text{if } i= target\:vehicle\\
    \hat{\Pi}_j^i(s_t),& \text{otherwise}
\end{cases}
\]
where $D_j$ is the maneuver model of the maneuver type $j$ (left lane change, right lane change, accelerate, decelerate, maintain), $s_{t_j}^i$ is the state of vehicle $i$ when the maneuver $j$ started. 
So essentially, decision variables here are j for $i=target\:vehicle$. 
\paragraph{Computing level-k policies $\Pi_k$:}
Let's consider 2 agents A and B. Agent A is trying to come up with a level-2 policy i.e. it needs to assume that agent B follows level-1 policy. Agent B follows level-1 policy by assuming level-0 policy of agent A. We find the level-0 policy of agent A by assuming that all vehicles would do maintain maneuver. We maximize  eq. [2] and obtain level-0 policy for agent A. Next, we use this level-0 policy of agent A and maximize eq. [2] to obtain level-1 policy for agent B. We keep repeating this to obtain level-0, level-1 and level-2 policies for agent B. The final step is to use eq. [3] and the probability model $P$ to obtain the best probable policy on an average.

\paragraph{Updating reasoning model $P$:}
We update the probabilities $P(k=i)$ as real data comes in at every time step as follows
\[
k^*=\arg\min_k (s_t(k)-s^m_t) \:\:\:\:\:\dots[6]
\]
\[
P(k=k^*)=P(k=k^*)+\delta \:\:\:\:\:\dots[7]
\] 
\[
P(k=i)=\frac{P(k=i)}{\sum_{k=0}^2P(k)} \:\:\:\:\:\dots[8]
\]
where $\delta$ is a fixed update step, $s_t(k)$ is the simulated joint state at time t if the vehicle being modeled had level-k policy, and $s^m_t$ is the measured joint state.  

\begin{minipage}{12.1cm}

\begin{algorithm}[H]
\caption{Lead Vehicle Maneuver Prediction}
\begin{algorithmic}[1]
\State \textbf{Input}: $State_{measured}:[x_{relative}(t),y_{relative}(t),v_x(t),v_y(t)]$, T, P
\State /// Level-0 policy for vehicle i is `No Action` 
\State $a^i_0=NO\_ACTION\:\forall i\in[0\isep m]$
\State
\State /// Solve for Level-0 policy of each vehicle i (m optimization problems) 
\State $\Pi^i_0=\arg\max_{a^i_{t'}}\sum_{t'=t}^{t+t_{maneuver(i, t)}} \gamma^{t'-t} C(s_t|a^j_0)\:\forall j\in[1\isep,m]-\{j=i\}$
\State
\State /// Evaluate level-0, level-1, level-2 policies of non-targets
\For {i = 1, 2, \dots, m}
\For {k = 1, 2}
\State $\hat{\Pi}^i_k=\arg\max_{a^i_{t'}}\sum_{t'=t}^{t+t_{maneuver(i, t)}} \gamma^{t'-t} C(s_t|\hat{\Pi}^j_{k-1})\: \forall j\in[1\isep,m]-\{j=i\}$
\EndFor
\EndFor
\State
\State /// Estimate policy of the target vehicle 
\State $\Pi=\arg\max_{a_{t'}} \sum_{k=0}^{2} P(k) \sum_{t'=t}^{t+t_{maneuver(i,t)}}C(\::|\:\hat{\Pi}_k^i)    \: \forall i\in[1\isep m]-\{i=target\:vehicle\}$
\State
\State ///Update P(k) using [6][7][8]
\State
\State \textbf{Return} $\Pi$
\end{algorithmic}
\end{algorithm}
\end{minipage}
\newline
\newline
\noindent where $t_{maneuver(i, t)}$ is the time to complete certain maneuver by vehicle i if the maneuver start time is t.

\section{Experiments on NGSIM data}
Initial validation of the algorithm was done on a dataset derived from NGSIM data by further processing it and breaking it down on per vehicle basis and bundling data frames relevant to each vehicle. 

\subsection{Velocity Resolution and Filtering}
To enrich the raw trajectory data with kinematically meaningful descriptors, we augmented each record with longitudinal and lateral velocity components. Let $x(t)$ and $y(t)$ denote the longitudinal (along-lane) and lateral (cross-lane) positions after projecting global coordinates into the road-aligned Frenet (or locally tangent) frame. The component-wise velocities were obtained via finite differences, then low-pass filtered to attenuate sensor and discretization noise. Filtering ensures smooth derivatives for downstream learning and reduces spurious accelerations that can bias a margin-based classifier like Support Vector Machine (SVM) which we'll be using later on for maneuver prediction. Savitzky–Golay filter was used and the exact parameters were chosen to preserve typical maneuver time scales while suppressing frame-to-frame jitter.

\subsection{Ego Vehicle Neighborhood Definition}
For every frame associated with a particular vehicle, a fixed rectangular spatial neighborhood centered on each “ego” vehicle was considered. At each time step, we selected as neighbors all vehicles whose positions fell into this rectangle i.e. within a window of 150 ft (45.72 m) longitudinal by 30 ft (9.144 m) lateral relative to the ego vehicle. This footprint captures the most influential vehicles in typical multi-lane traffic: leaders and followers within several car lengths and immediate left/right lane vehicles. The rectangle was applied in the road-aligned frame to avoid distortion from roadway curvature; bounds were symmetric around the ego vehicle.

\subsection{Per-Vehicle Data Packaging}
For each ego vehicle, we constructed a self-contained sample file comprising:
\begin{itemize}
    \item the ego vehicle’s time-synchronized states (position, filtered $velocity_{longitudinal}$, filtered $velocity_{lateral}$, and any available headings/accelerations).
    \item the corresponding states of all neighbors identified by the window at each timestamp.
\end{itemize}

Packaging at the per-ego granularity simplifies parallel processing, eliminates unnecessary computational overhead and simplifies train/validation splits for the aforementioned algorithm. Files were serialized in a consistent schema (timestamps, ego features, followed by a variable-length list of neighbor features with IDs), enabling batched loading. When no neighbors were present, the record included an explicit empty set to avoid sampling bias.

\subsection{Evaluation}
Algorithm 1 was implemented in MATLAB (referred to as interaction-based predictor hereafter) and the processed dataset was fed in as a sequence of frames per ego vehicle to evaluate the combined performance of the motion-based predictor and interaction-based predictor.
We selected a Support Vector Machine (SVM) as the motion-based predictor. SVMs provide strong performance with modest dataset sizes, are robust to high-dimensional yet correlated features (e.g., relative positions and velocities), and yield decent classification performance for short-horizon maneuver classification or labeling. Prior to training, all continuous inputs were standardized (zero mean, unit variance). Kernel choice and regularization were tuned via cross-validation to balance margin maximization against overfitting.
The SVM-based motion predictor was first used solely to establish a baseline on predictor performance against the ground truth i.e. the trajectories and maneuvers actually done by the vehicles surrounding the ego vehicle in the dataset. The average prediction accuracy of the validation set came out to be 73\%. The misclassifications were due to over-reliance on motion cues and no interaction-based reasoning.   
Then the interaction-based predictor was evaluated solely to compare against the baseline. Since it didn't rely on motion cues, it could make predictions with high confidence early on but suffered from driver modeling inaccuracies which pulled down the average classification performance.
The combination of both SVM-based motion predictor and the interaction-based predictor utilizing the level-k reasoning outperformed the baseline significantly by averaging the maneuver prediction accuracy to 90\%. It also predicted mostly correct maneuvers over longer horizons which resulted in predicting complex behaviors like double lane changes.
\begin{figure}[h]
\centering
\includegraphics[width=0.6\textwidth]{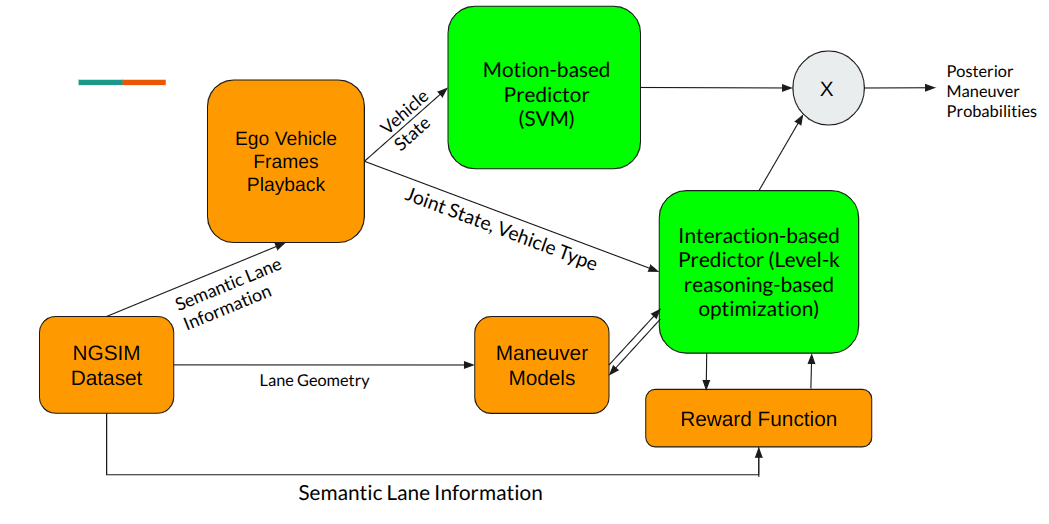}
\caption{System Architecture}
\end{figure}

\subsection{Challenges}
There is an unintended bias introduced during NGSIM dataset processing due to labeling criteria of lane changes in the trajectories. Another factor that greatly affects the accuracy of the interaction-based predictor and the combined approach is the initial driver level estimates and update scheme. Slower update rates can lead to lower confidence in the predictions and faster updates could lead to misclassifications. The optimization process gets computationally expensive very quickly as the number of participants surrounding the ego vehicle increase and also with an increase in the level-k reasoning level. Thus optimization at algorithm implementation level is required to maintain feasibility.

\subsection{Conclusion}
In spite of the additional computation due to modeling of the interactions between different vehicles in the scene, a significant improvement in the maneuver prediction accuracy proves that incorporating social interactions between agents is essential to achieve reasonable real-world usability of a prediction module. Data-driven conclusions can be used to reduce the computational overhead by limiting the k-levels to {0,1}. Better driver modeling approaches can be used to learn the level of reasoning of an agent using minimum data. This work is a step in direction of developing more integrated behavior prediction solutions which bring together data-driven and multi-agent reasoning.     

\bibliographystyle{plain}

\end{document}